\documentclass[10pt,twocolumn,letterpaper]{article}

\usepackage{iccv}
\usepackage{times}
\usepackage{epsfig}
\usepackage{graphicx}
\usepackage{amsmath}
\usepackage{amssymb}
\usepackage{multirow}
\usepackage{makecell}
\usepackage{booktabs}
\usepackage{colortbl}
\usepackage{xcolor}
\usepackage{mathtools}
\usepackage{array}
\usepackage{pifont}

\newcommand{\baseline}[1]{\textit{\textcolor{gray}{#1}}}
\newcommand{\increment}[1]{\small {\color{purple}{#1}}} 
\newlength\savewidth\newcommand\shline{\noalign{\global\savewidth\arrayrulewidth
  \global\arrayrulewidth 1pt}\hline\noalign{\global\arrayrulewidth\savewidth}}
\usepackage[pagebackref=true,breaklinks=true,letterpaper=true,colorlinks,bookmarks=false]{hyperref}

\iccvfinalcopy 

\ificcvfinal\fi

\begin{document}

\title{From Knowledge Distillation to Self-Knowledge Distillation: A Unified Approach with Normalized Loss and Customized Soft Labels}

\author{Zhendong Yang$^{1,2}$\quad Ailing Zeng$^{2}$ \quad Zhe Li$^{3}$\quad Tianke Zhang$^{1}$
\quad Chun Yuan$^{1*}$ \quad Yu Li$^{2*}$\\
$^{1}$Tsinghua Shenzhen International Graduate School
\quad$^{2}$International Digital Economy Academy (IDEA)\\ $^{3}$Institute of Automation, Chinese Academy of Sciences\\
{\tt\small \{yangzd21,ztk21\}@mails.tsinghua.edu.cn \quad axel.li@outlook.com}\\
{\tt\small yuanc@sz.tsinghua.edu.cn \quad \{zengailing, liyu\}@idea.edu.cn}
}

\maketitle
\ificcvfinal\thispagestyle{empty}\fi

\renewcommand{\thefootnote}{\fnsymbol{footnote}} 
\footnotetext[1]{Corresponding authors} 

\begin{abstract}
   Knowledge Distillation (KD) uses the teacher's prediction logits as soft labels to guide the student, while self-KD does not need a real teacher to require the soft labels. This work unifies the formulations of the two tasks by decomposing and reorganizing the generic KD loss into a Normalized KD (NKD) loss and customized soft labels for both target class (image's category) and non-target classes named Universal Self-Knowledge Distillation (USKD). We decompose the KD loss and find the non-target loss from it forces the student's non-target logits to match the teacher's, but the sum of the two non-target logits is different, preventing them from being identical. NKD normalizes the non-target logits to equalize their sum. It can be generally used for KD and self-KD to better use the soft labels for distillation loss. USKD generates customized soft labels for both target and non-target classes without a teacher. It smooths the target logit of the student as the soft target label and uses the rank of the intermediate feature to generate the soft non-target labels with Zipf's law. For KD with teachers, our NKD achieves state-of-the-art performance on CIFAR-100 and ImageNet datasets, boosting the ImageNet Top-1 accuracy of ResNet18 from 69.90\% to 71.96\% with a ResNet-34 teacher. For self-KD without teachers, USKD is the first self-KD method that can be effectively applied to both CNN and ViT models with negligible additional time and memory cost, resulting in new state-of-the-art results, such as 1.17\% and 0.55\% accuracy gains on ImageNet for MobileNet and DeiT-Tiny, respectively. Our codes are available at \url{https://github.com/yzd-v/cls_KD}.
\end{abstract}

\section{Introduction}

\begin{figure}[t]
    \centering
    \includegraphics[width=\linewidth]{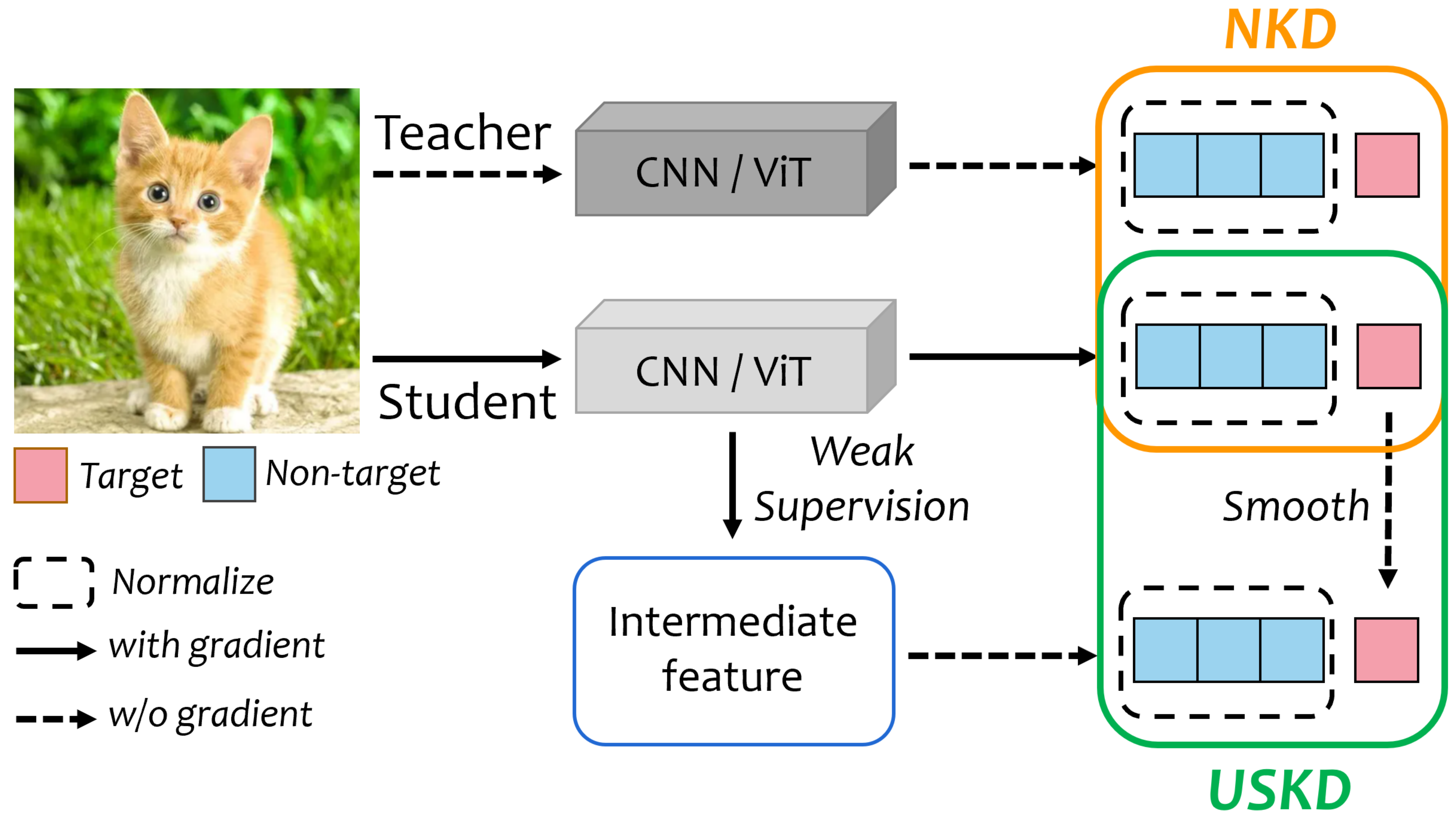}
    \caption{Illustration of the proposed NKD and USKD for distillation loss calculations. NKD normalizes the non-target logits, using the soft labels more effectively, and achieves better performance. Meanwhile, USKD sets customized soft labels for both target and non-target classes, and can be applied to both CNNs and ViTs.}
    \label{fig:all}
\end{figure}

Deep convolutional neural networks (CNNs) have significantly advanced the performance in many tasks~\cite{he2017mask,he2016deep,ren2015faster,ronneberger2015u}. In general, a larger model performing better needs more computing resources. On the other hand, smaller models have lower computation complexity but are less competitive than larger models. To bridge this gap and improve the performance of smaller models, knowledge distillation (KD) has been proposed~\cite{hinton2015distilling}. The core idea of KD is to employ the teacher's prediction logits as soft labels to guide the student. Self-knowledge distillation (self-KD)~\cite{sun2019deeply,zhang2019your} is inspired by the knowledge distillation method, but it does not require an actual teacher. Instead, it designs soft labels through auxiliary branches or special distribution. The similarity between KD and self-KD is that they utilize soft labels for distillation loss, while the key difference is in how they obtain the soft labels. This paper aims to 1) improve the utilization of soft labels for distillation loss and 2) propose a general and effective method to obtain customized soft labels for self-KD. The targets make us obtain the soft labels with a teacher and use our modified distillation loss for better performance. Alternatively, when we lack a teacher, we can use the proposed self-KD method to obtain the soft labels and then calculate the loss.

The original cross-entropy (CE) loss for classification calculates the loss on the target class (the image's category).
While the soft labels from the teacher include target and non-target class, thus the KD loss also includes both target and non-target loss. The decoupled method has been proven effective for KD in DKD~\cite{zhao2022decoupled}. 
Unlike DKD's method of adjusting hyper-parameters on target and non-target loss, we present a simple yet effective way to decompose the KD loss. We decompose the KD loss into a combination of the target loss (like the original CE loss) and the non-target loss in CE form. The non-target loss transforms the internal distribution of the student's non-target logits to match the teacher's distribution. However, we find that the sum of the student's and teacher's non-target logits is changing and different, which hinders the alignment of their distributions. To address this issue, we normalize the non-target logits to equalize their sum, transferring teacher's non-target knowledge. With this slight modification, we introduce our Normalized Knowledge Distillation (NKD) loss, as depicted in Fig.\ref{fig:all}, significantly enhancing KD's performance.

\begin{figure*}[t]
    \centering
    \includegraphics[width=1.0\linewidth]{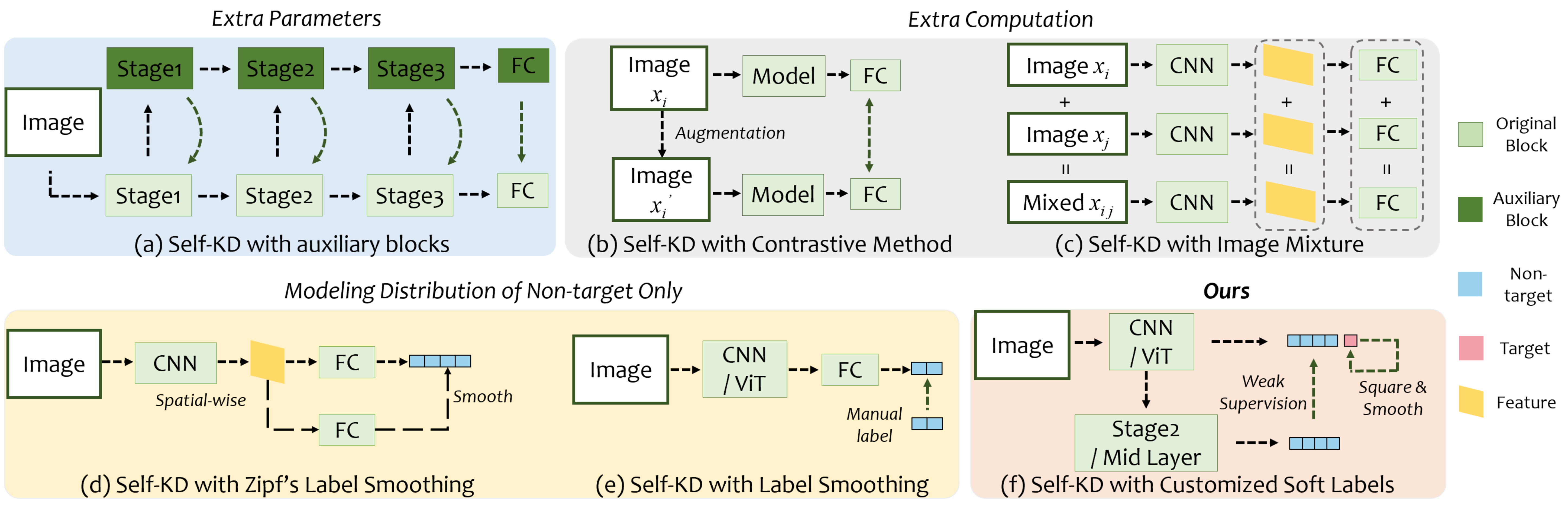}
    \caption{Comparison of different self-KD methods with \textbf{(a)} \emph{extra blocks and parameters}~\cite{ji2021refine}, \textbf{(b)} \emph{contrastive method}~\cite{xu2019data}, \textbf{(c)} \emph{image mixture}~\cite{yang2022mixskd},  \textbf{(d)} \emph{Zipf's label smoothing}~\cite{liang2022efficient}, and \textbf{(e)} \emph{label smoothing}~\cite{muller2019does,szegedy2016rethinking}. \textbf{(f)} Self-KD with our \emph{customized soft labels} including both target and non-target class for distillation. Note some methods can only be applied to CNN, while ours works for both CNN and ViT. }
    \label{fig:skd methods}
\end{figure*}

Our proposed NKD utilizes the teacher's target logit and normalized non-target logits to guide the student, resulting in state-of-the-art performance. This demonstrates the effectiveness of NKD loss formulation. Also, it can be generally used for self-KD to calculate the distillation loss, but how to generate the soft labels without a real teacher generally and efficiently is also important.

Various self-KD methods have explored using manually designed soft labels to enhance students with less time than KD. These methods~\cite{ji2021refine,yang2022mixskd,yun2020regularizing,zhang2019your} typically obtain the labels from auxiliary branches or contrastive learning, as depicted in Fig.~\ref{fig:skd methods} (a), (b), and (c). However, despite requiring less time than KD, they still involve significant overhead compared to training the model directly. Recently, state-of-the-art Zipf's LS~\cite{liang2022efficient}, as shown in Fig.~\ref{fig:skd methods} (d), introduced soft non-target labels based on a special distribution that can significantly reduce resource and time requirements. It classifies the student's feature in the spatial dimension and determines the rank of the non-target class using Zipf's law~\cite{Powers1998ApplicationsAE}. However, it requires the pixel-level features before average pooling, making it unsuitable for ViT-like models~\cite{dosovitskiy2021an} with patch-wise tokenization.

To address the limitations of existing methods, we propose a general and effective way to obtain soft labels. We design customized soft labels available for both CNN and ViT models. Following the NKD loss formulation, our customized soft labels comprise soft target label and soft non-target labels for corresponding loss. For the soft target label, we replace the teacher's target logit with the smoother label value obtained from the student's prediction. Since the student's predictions vary drastically during training, especially in the beginning, we smooth the student's target output within each training batch to stabilize the label values. For the soft non-target labels, we need their rank and distribution. First, for the rank, we get it from the intermediate feature, making our method available for both CNN and ViT models. We take weak supervision on the intermediate feature to get weak logit. Then, we normalize and combine it with the final logit and sort for the rank, as shown in Fig.~\ref{fig:skd methods} (f). The soft non-target labels' distribution follows Zipf's Law~\cite{Powers1998ApplicationsAE}. With the soft target and non-target labels, we set our customized soft labels and propose Universal Self-Knowledge Distillation (USKD) as shown in Fig.~\ref{fig:all}. Besides, USKD only needs an extra linear layer for weak supervision. So it just takes a few more computing resources and time than training the model directly. USKD is a simple and effective method that achieves state-of-the-art performance on both CNN and ViT models.

As described above, we normalize KD's non-target logits and propose NKD, using the soft labels better and improving KD's performance significantly. For the generation of soft labels without a real teacher, we set soft target and non-target labels, proposing USKD for self-KD. In a nutshell, the contributions of this paper are:
\begin{itemize}
  \item
  We normalize the non-target logits in the classical KD, making it better to optimize the cross-entropy loss. With this minor change, we propose Normalized KD (NKD) loss, using teacher's soft labels better and improving KD's performance significantly.
  \item
  We propose a novel way to set customized soft labels without a real teacher, including target and non-target classes for self-KD. We utilize the weak logit to obtain soft non-target labels. Besides, we enlarge the difference between student's target logit for different images and soften them for soft target labels.
  \item
  We propose a simple and effective self-KD method USKD with our customized soft labels, which applies to both CNN and ViT models. Importantly, USKD requires only almost negligible additional time and resources compared to training the model directly.
  \item
  We conduct extensive experiments on CIFAR-100 and ImageNet to verify the effectiveness of NKD and USKD, achieving state-of-the-art performance. Additionally, we demonstrate the efficacy of models trained with our self-KD method on COCO for detection.
\end{itemize}

\section{Related work}
\subsection{Knowledge Distillation}
Knowledge distillation is a method to improve the model while keeping the network unchanged. It was first proposed by Hinton~\etal~\cite{hinton2015distilling}, where the student is supervised by the hard and soft labels from the teacher's output. Many following works focus on better using soft labels to transfer more knowledge. WSLD~\cite{zhou2020rethinking} analyzes soft labels and distributes different weights for them from a perspective of bias-variance trade-off. SRRL~\cite{yang2020knowledge} forces the output logits of the teacher's and student's features after the teacher's linear layer to be the same. DKD~\cite{zhao2022decoupled} decouples the logit and distributes different weights for the target and non-target classes. DIST~\cite{huang2022knowledge} uses the Pearson correlation coefficient to replace the KL divergence and transfers the inter-relation and intra-relation together.

Besides distillation on logits, some works~\cite{cao2022pkd,shu2021channel,yang2022focal,yang2022vitkd} aim at transferring knowledge from intermediate features. FitNet~\cite{romero2014fitnets} distill the semantic information from the intermediate feature directly. AT~\cite{zagoruyko2016paying} transfers the attention of feature maps to the students. OFD~\cite{heo2019comprehensive} designs the margin ReLU and modifies the measurement for the distance between students and teachers. RKD~\cite{park2019relational} extracts the relation from the feature map. CRD~\cite{tian2019contrastive} applies contrastive learning to distillation successfully. KR~\cite{chen2021distilling} transfers knowledge from multi-level features for distillation. TaT~\cite{lin2022knowledge} helps the student to learn the teacher's every spatial component. MGD~\cite{yang2022masked} masks the student's feature and forces it to generate the teacher's feature.

\subsection{Self-Knowledge Distillation}

Self-Knowledge Distillation has been proposed as an alternative approach to Knowledge Distillation that does not rely on an external teacher model. Self-KD aims to utilize the information within the student model to guide its learning process. Several self-KD methods have been proposed in recent years. DKS~\cite{sun2019deeply} introduces auxiliary supervision branches and pairwise knowledge alignments, while BYOT~\cite{zhang2019your} adds blocks and layers to every stage and uses shallow and deep features as student and teacher, respectively. KDCL~\cite{guo2020online} trains two models for online knowledge distillation, while FRSKD~\cite{ji2021refine} adds a new branch supervised by the original feature and uses the logit of the new branch for self-KD. DDGSD~\cite{xu2019data} transfers knowledge between different distorted versions of the same training data. OLS~\cite{zhang2021delving} sets a matrix that is made up of the soft label for every class, while MixSKD~\cite{yang2022mixskd} proposes incorporating self-knowledge distillation with image mixture and aggregates multi-stage feature to produce soft labels. Tf-FD~\cite{li2022self} includes intra-layer and inter-layer distillation, reusing the channel-wise and layer-wise features to provide knowledge without an additional model. However, these methods require auxiliary architecture, adapt layers for alignment, or data augmentation, consuming much more time and computing resources than training the model directly.

In contrast, some self-KD methods require little extra or even no more time than training the model directly. For example, Label Smoothing~\cite{szegedy2016rethinking} sets the labels manually by distributing the same values to all non-target classes. Tf-KD~\cite{yuan2020revisiting} revisits KD via label smoothing, using a high temperature to generate the manual logit for distillation. Zipf's LS~\cite{liang2022efficient} utilizes the student's linear layer to obtain several logits for each pixel of the student's last feature map. The method uses these logits to vote for the non-target class's rank with Zipf distribution~\cite{Powers1998ApplicationsAE}. These methods set soft labels to achieve self-knowledge distillation without contrastive learning, data augmentation, or auxiliary branches, saving much training time and computing resources.

\section{Method}
\label{method}
\subsection{Normalized KD for Better Using Soft Labels}
\label{kd with t}

Using $t$ denote the target class, $C$ denote the number of classes, $V_{i}$ denote the label value for each class $i$, and $S_{i}$ denote the student's output probability. The original loss for image classification can be formulated as follows:
\begin{align}
    L_{ori} = -\sum_{i}^{C}V_{i}log (S_{i})=-V_{t}log ( S_{t})=-log (S_{t}).
\end{align}
Using $T_{i}$ denote the teacher's soft labels. The classical KD utilizes them for distillation loss as:
\begin{align}
\label{eq:original kd begin}
    L_{kd}&=-\sum_{i}^{C} T_{i}log (S_{i})\\
    &=-T_{t}log (S_{t}) - \sum_{i \ne t}^{C}T_{i}log(S_{i}).
\end{align}
As $L_{kd}$ shows, the first loss $-T_{t}log (S_{t})$ is about the target class like the original $L_{ori}$. While for the second non-target loss $-\sum_{i \ne t}^{C}T_{i}log (S_{i})$, it has the same form as CE loss $-\sum p (x)log (q (x))$. The CE loss aims at making $q (x)$ be the same as $p (x)$. However, $\sum_{i\ne t}^{C} T_{i}= 1-T_{t}$ and $\sum_{i\ne t}^{C}S_{i}= 1-S_{t}$. Since the trainable $S_{i}$ is unlikely to exactly match the fixed $T_{i}$ during training, the sum of the two non-target logits is always different, preventing the two distributions from being the same. Thus we normalize $T_{i}$ and $S_{i}$ to force the sum of the two distributions to be the same. With $\mathcal{N}(\cdot)$ denoting the normalized operation, we modify the KD loss and propose our \textbf{N}ormalized \textbf{K}nowledge \textbf{D}istillation (NKD) loss as follows:
\begin{align}
\label{eq:new kd}
    L_{nkd}=-T_{t}log (S_{t}) - \gamma\cdot \lambda^{2}\cdot\sum_{i \ne t}^{C}\mathcal{N}(T_{i}^{\lambda})log(\mathcal{N}(S_{i}^{\lambda})),
\end{align}
where $\gamma$ is a hyper-parameter to balance the loss and $\lambda$ is the temperature for KD~\cite{hinton2015distilling}. Finally, combining the original loss $L_{ori}$, and NKD loss $L_{nkd}$, we train the students with:
\begin{equation}
    \label{eq:all loss}
        L_{all}=L_{ori}+ L_{nkd}.
\end{equation}

\subsection{Customized Soft Labels for Self-KD}
We utilize NKD for better use of the soft labels. Following our NKD loss in Eq.~\ref{eq:new kd}, we propose a general and effective self-KD method, which sets soft labels on target and non-target classes without teachers.

\subsubsection{Soft Target Label}
First, for the soft target label, the weight $T_{t}$ in Eq.~\ref{eq:new kd} is obtained from the teacher's target output probability for the input image. We wonder if the soft target can be provided by adjusting the student's target output $S_{t}$. The differences between $T_{t}$ and $S_{t}$ mainly focus on two parts. The first is that $T_{t}$ is fixed, and $S_{t}$ varies gradually during training. The second part is that the difference between different samples' $S_{t}$ is much smaller than $T_{t}$ at the beginning of training. To overcome the problem, we first square $S_{t}$ to enlarge the difference between different samples' $S_{t}$. Then we propose a way to adjust $S_{t}$, making it smoother to fit the training set without teachers. This strategy can be applied directly to different models, including CNN-liked and ViT-liked models. In this way, we get the soft target label $P_{t}$, which is detached and zero-grad in training for self-KD:
\begin{align}
\label{eq:smooth}
    P_{t} = S_{t}^{2}+V_{t}-mean (S_{t}^{2}).
\end{align}
With the $P_{t}$, we follow our NKD for the target loss:
\begin{align}
\label{eq:target loss}
    L_{target} = - P_{t}log (S_{t}),
\end{align}
where $V_{t}$ denote the original target label value, \emph{e.g.} [0.8,0.2] for a mixed image. And $mean(\cdot)$ is calculated across different samples in a training batch. We discuss the effects of the smoothing ways for $S_{t}$ in Sec.~\ref{sec:different smooth}. 

\subsubsection{Soft Non-target Labels}

The knowledge from the teacher's soft non-target labels includes its rank and distribution. We also need them to set the soft non-target labels for self-KD. To get the rank, we first obtain a new weak logit by setting weak supervision on the intermediate feature. Using $\mathcal{F}$ denote the feature of stage 2 of the CNN-liked model or the mid layer's classification token of the ViT-liked model, $GAP$ denote the global average pooling, $FC$ denote a new linear layer, the weak logit of the CNN-liked model can be formulated as:
\begin{align}
\label{eq:weak}
    W_{i} = softmax(FC(GAP(\mathcal{F}))),
\end{align}
As for the ViT-liked model, the weak logit is as follows:
\begin{align*}
    W_{i} = softmax(FC(\mathcal{F})).
\end{align*}

We aim to obtain another smoother non-target logit different from the original final logit. So we utilize a smooth label and take weak supervision to get the weak logit. Using $V_{i}$ denote the label, which is the original label processed with label smoothing, the loss for obtaining the weak logit is as follows:
\begin{align}
\label{eq:weak loss}
    L_{weak} =-\mu\cdot \sum_{i}^{C} V_{i}log (W_{i}),
\end{align}
where $\mu<1$ is a hyper-parameter to achieve weak supervision. With the weak logit, we obtain the soft non-target labels' rank by normalizing and combining it with the final logit. This operation balances the two logits' effects for the rank, and the analysis is shown in Sec.~\ref{sec:normal}.
\begin{align}
\label{eq:non-target sort}
    R_{i} = \frac{W_{i}}{1-W_{t}} + \frac{S_{i}}{1-S_{t}}.
\end{align}

The soft non-target labels' distribution follows Zipf's law~\cite{Powers1998ApplicationsAE}, which has been applied in Zipf's LS~\cite{liang2022efficient}. The formulation can be formulated as follows:
\begin{align}
\label{eq:zipf}
    Z_{i} = \frac{i^{-1}}{\sum_{i=1}^{C} i^{-1}}.
\end{align}
With the distribution, we sort it with the rank of $R_{i}$ and obtain the soft non-target labels $Z_{i}$ for self-KD. Following our NKD loss in Eq.~\ref{eq:new kd}, the non-target loss is as follows:
\begin{align}
\label{eq:non loss}
    L_{non} = - \sum_{i\ne t}^{C}\mathcal{N}(Z_{i})log (\mathcal{N}(S_{i})).
\end{align}

\subsubsection{Overall for Self-KD}
With the proposed soft target label and soft non-target labels, we calculate the corresponding loss and propose USKD as shown in Fig.~\ref{fig:all}. We train all the models with the total loss for self-KD as follows:
\begin{equation}
    \label{eq:all loss}
        L_{all}=L_{ori}+\alpha\cdot L_{target} + \beta\cdot L_{non} + L_{weak},
\end{equation}
where $L_{ori}$ is the original loss for the models among all the tasks, $\alpha$ and $\beta$ are two hyper-parameters to balance the loss scale. $L_{weak}$ is used to generate the weak logit in Eq.~\ref{eq:weak}.

\section{Experiments}
\label{main experiments}

\subsection{Datasets and Details}
We conduct the experiments on CIFAR-100~\cite{krizhevsky2009learning} and ImageNet~\cite{deng2009imagenet}, which contain 100 and 1000 categories, respectively. For CIFAR-100, we use 50k images for training and 10k for validation. For ImageNet, we use 1.2 million images for training and 50k images for validation. In this paper, we use accuracy to evaluate all the models.

For KD with teachers, NKD has two hyper-parameters $\gamma$ and $\lambda$ in Eq.~\ref{eq:new kd}. For all the experiments, we adopt $\{\gamma = 1.5, \lambda=1\}$ on ImageNet. While for CIFAR-100, we follow the training setting from DKD~\cite{zhao2022decoupled} for a fair comparison. And USKD has two hyper-parameters $\alpha$ and $\beta$ to balance the loss scale in Eq.~\ref{eq:all loss}. Another hyper-parameter $\mu$ is used to achieve the weak supervision in Eq.~\ref{eq:weak loss}. For all the experiments, we adopt $\{\alpha = 1, \beta = 0.1, \mu=0.005\}$ on ImageNet and $\{\alpha = 0.1, \beta = 0.1, \mu=0.1\}$ on CIFAR-100. The other training setting for KD and self-KD is the same as training the students without distillation. We use 8 GPUs to conduct the experiments with MMClassition~\cite{2020mmclassification} based on Pytorch~\cite{paszke2019pytorch}. More details and experimental results about the hyper-parameters are shown in the supplement.

\subsection{Normalized KD with Teachers}

\begin{table*}[t]
\begin{center}
\begin{tabular}{c|c|cccc|cc}
\shline
Model&\makecell{Teacher\\Student} & \makecell{VGGNet13\\VGGNet8} &\makecell{ResNet32x4\\ResNet8x4}&\makecell{VGGNet13\\MobileNetV2}&\makecell{ResNet50\\MobileNetV2}&\makecell{ResNet34\\ResNet18}&\makecell{ResNet50\\MobileNet}\\
\shline
\multirow{2}{*}{\makecell{Accuracy}}
&Teacher & 74.64 & 79.42 & 74.64 & 79.34 & 73.62 & 76.55\\
&\baseline{Student} & \baseline{70.36}& \baseline{72.50}& \baseline{64.60}&\baseline{64.60}&\baseline{69.90}&\baseline{69.21}\\
\hline
\multirow{4}{*}{\makecell{Feature}}
&RKD~\cite{park2019relational}&71.48&71.90&64.52&64.43&71.34&71.32\\
&CRD~\cite{tian2019contrastive}&73.94&75.51&69.73&69.11&71.17&71.40\\
&OFD~\cite{heo2019comprehensive}&73.95&74.95&69.48&69.04&71.08&71.25\\
&KR~\cite{chen2021distilling}&\underline{74.84}&75.63&{\bf70.37}&69.89&71.61&\underline{72.56}\\
\hline
\multirow{4}{*}{\makecell{Logit}}
&KD~\cite{hinton2015distilling}&72.98&73.33&67.37&67.35&71.03&70.68\\
&WSLD~\cite{zhou2020rethinking}&74.36&76.05&69.02&70.15&\underline{71.73}&72.02\\
&DKD~\cite{zhao2022decoupled}&74.68&\underline{76.32}&69.71&\underline{70.35}&71.70&72.05\\
&{\bf NKD (ours)}&{\bf74.86}&{\bf76.35}&\underline{70.22}&{\bf70.67}&{\bf71.96}&{\bf72.58}\\
\shline
\end{tabular}
 \end{center}
\caption{Results of different knowledge distillation methods on CIFAR-100 (the left four columns) and ImageNet (the right two columns) dataset. The data that is underlined denotes the second-best result among all the results. The metric is the Top-1 accuracy (\%).}
\label{table:cifar results}
\end{table*}

\begin{table*}
  \begin{minipage}[t]{0.53\linewidth}
  \vspace{0pt}
  \setlength{\tabcolsep}{3.6 pt}
  \begin{center}
  \begin{tabular}{l|c|ccc}
    \shline
    Dataset & CIFAR100 & \multicolumn{3}{c}{ImageNet}\\
    \hline
    Model   & ResNet18 & ResNet18 & ResNet50 & MobileNet\\
    \shline
    \baseline{Baseline}      & \baseline{78.58} & \baseline{69.90} & \baseline{76.55} & \baseline{69.21}\\
    \hline
    LS~\cite{szegedy2016rethinking}    &79.42 &69.92&76.64&68.98\\
    Tf-KD~\cite{yuan2020revisiting}   &79.53 &70.14&76.59&69.20\\
    Zipf's LS~\cite{liang2022efficient}  &79.63&70.30&76.96&69.59 \\
    {\bf USKD (ours)}     &{\bf79.90}&{\bf70.79}&{\bf77.07}&{\bf70.38}\\
    \shline
  \end{tabular}
   \end{center}
    \caption{The comparative results of different self-knowledge distillation methods on CIFAR100 and ImageNet dataset. We report the models' performance with Top-1 accuracy (\%).}
  \label{table:main results}
  \end{minipage}
  \hspace{2pt}
  \begin{minipage}[t]{0.44\linewidth}
  \vspace{0pt}
   \setlength{\tabcolsep}{3.6 pt}
   \renewcommand\arraystretch{1.012}
   \begin{center}
  \begin{tabular}{c|ccc}
    \shline
    &Model   & \baseline{Baseline}   &{\bf USKD (ours)} \\
    \hline
    \multirow{3}{*}{\makecell{Accuracy\\(\%)}}
    &ResNet-101   & \baseline{77.97} & 78.54 (\increment{+0.57})\\
    &MobileNet-V2  & \baseline{71.86} & 72.41 (\increment{+0.55})\\
    &ShuffleNet-V2  & \baseline{69.55} & 70.30 (\increment{+0.75})\\
    \hline
    \multirow{3}{*}{\makecell{Time\\(min)}}
    &ResNet-101   & \baseline{13.78} & 13.95 \\
    &MobileNet-V2  & \baseline{10.17} & 10.18\\
    &ShuffleNet-V2  & \baseline{8.63} & \ \ 8.68 \\
    \shline
  \end{tabular}
   \end{center}
  \caption{ The results of the models' performance are Top-1 accuracy (\%) on ImageNet dataset. The time data are reported with minutes (min) for a training epoch.}
  \label{table:more comparison}
  \end{minipage}
\end{table*}

When we get the soft labels from a real teacher, we can use NKD loss for better performance. To prove this, we first conduct experiments with various teacher-student distillation pairs on CIFAR-100, shown in Tab.~\ref{table:cifar results}. In this setting, we evaluate our method on several models with different architectures, including VGGNet~\cite{simonyan2014very}, ResNet~\cite{he2016deep}, ShuffleNet~\cite{zhang2018shufflenet} and MobileNetV2~\cite{sandler2018mobilenetv2}. We compare our method with KD~\cite{hinton2015distilling} and several other state-of-the-art distillation methods. As the results show, our method brings the students remarkable accuracy gains over other methods. Our method achieves the best performance among logit-based distillation methods and even surpasses the feature-based distillation methods in some settings.

To further demonstrate the effectiveness and robustness of our NKD, we test it on a more challenging dataset, ImageNet. We set two popular teacher-student pairs, which include homogeneous and heterogeneous teacher-student structures for distillation. The homogeneous distillation is ResNet34-ResNet18, and the heterogeneous distillation is ResNet50-MobileNet. The results of different KD methods on ImageNet are shown in Tab.~\ref{table:cifar results}. As the results show, our method outperforms all the previous methods. It brings consistent and significant improvements to the students for both distillation settings. The student ResNet18 and MobileNet achieve 71.96\% and 72.58\% Top-1 accuracy, getting 2.06\% and 3.37\% accuracy gains with the knowledge transferred from the teacher's logits, respectively.

As described above, our NKD enhances KD's performance significantly with a slight modification. And in various settings, it also surpasses DKD, a method that improves KD according to a different decoupled way.

\subsection{Universal Self-KD without Teachers}

When we lack a teacher, we use the proposed self-KD method USKD to obtain the soft labels
and then calculate the loss. To evaluate its effectiveness, we first conduct experiments with ResNets~\cite{he2016deep} and MobileNet~\cite{howard2017mobilenets} on CIFAR100 and ImageNet datasets. We compare with the other state-of-the-art methods, which also set manual labels and only bring little extra time consumption, including label smoothing~\cite{szegedy2016rethinking}, Tf-KD~\cite{yuan2020revisiting} and Zipf's LS~\cite{liang2022efficient}. As shown in Tab.~\ref{table:main results}, our method surpasses the previous related self-knowledge distillation methods on various settings and brings the model remarkable gains. For example, it brings MobileNet and ResNet-18 1.17\% and 0.89\% Top-1 accuracy gains on ImageNet.

Furthermore, we also test our USKD on more models, which include lighter models MobileNetV2~\cite{sandler2018mobilenetv2}, ShuffleNetV2~\cite{ma2018shufflenet} and a deeper ResNet. The results are shown in Tab.~\ref{table:more comparison}. For both the lightweight models, including MobileNetV2 and ShuffleNetV2, and the stronger model ResNet-101, our method also achieves considerable improvements. Besides, we also compare the time consumption to train the model for an epoch in Tab.~\ref{table:more comparison}. Compared with the baseline, the extra time we need for self-knowledge distillation is very limited. Specifically, the time of training MobileNet-V2 with our proposed USKD for an epoch is 10.18 minutes, which is just 0.01 minutes higher than training the model directly. Our method brings considerable improvements to various models with negligible extra time consumption. 

\begin{table}[t]
  \begin{center}
  \setlength{\tabcolsep}{10 pt}
  \begin{tabular}{lcc}
    \shline
    Model   &\baseline{Baseline}  &{\bf USKD  (ours)} \\
    \hline
    RegNetX-1.6GF   &\baseline{76.84}&77.30 (\increment{+0.46})\\
    DeiT-Tiny  &\baseline{74.42}&74.97 (\increment{+0.55}) \\
    DeiT-Small  &\baseline{80.55}&80.77 (\increment{+0.22}) \\
    Swin-Tiny  &\baseline{81.18}&81.49 (\increment{+0.31})\\
    \shline
  \end{tabular}
  \end{center}
\caption{Results of training more models, including ViT-liked models with our proposed method on ImageNet dataset. All the results are reported with Top-1 accuracy (\%).}
\label{table:vit}
\end{table}

\subsection{Universal Self-KD for More Models}
The previous self-KD methods are specially designed for CNN-liked models. However, some models like ViT~\cite{dosovitskiy2021an} translate the image into different tokens and have a completely different architecture. Those self-KD methods fail to benefit ViT-liked models. For ViT-liked models, USKD can also set customized soft labels and bring remarkable improvements. As described in Eq.~\ref{eq:weak}, we use an extra linear layer connected to ViT's middle layer for classification and obtain the weak logit. The rest operations are the same as CNN-liked models. To show the generalization of USKD, we apply it to more models, including RegNet~\cite{radosavovic2020designing}, DeiT~\cite{touvron2021training}, and Swin-Transformer~\cite{liu2021swin}, which can be seen in Tab.~\ref{table:vit}. All the models can achieve remarkable Top-1 accuracy gains. Besides, our method even can bring 0.55\% gains for DeiT-Tiny. And it also brings 0.31\% Top-1 accuracy gains for the latest state-of-the-art model Swin-Transformer. The results of more models show our method is both effective and general. 

\section{Analysis}

\subsection{Effects of Normalizing the Non-target Logits}
In this paper, we proposed normalizing the non-target logits in the original KD to help the student perform better. In this subsection, we conduct experiments to demonstrate the effectiveness of our modification. As shown in Tab.~\ref{table:ablation study}, using the target loss alone leads to a 1.16\% increase in accuracy. Combining the knowledge from both target and non-target classes allows us to better use the teacher's knowledge, resulting in a significant improvement of 2.06\% Top-1 accuracy for the student. Furthermore, we normalize the non-target logits in the KD loss for distillation and compare the non-target loss from both KD and NKD. Our non-target loss brings much greater gains than KD's non-target loss, as shown in the comparison. These results demonstrate the effectiveness of our proposed modification in improving KD's performance.
\begin{table}
  \begin{center}
  
  \begin{tabular}{ccccc}
    \shline
    Loss & \multicolumn{4}{c}{ResNet34 - ResNet18}\\
    \hline
    Target    & \ding{55}  &\ding{51}&\ding{51}&\ding{51}\\
    KD's Non-target   & \ding{55} &\ding{55}&\ding{51}&\ding{55}\\
    NKD's Non-target      &\ding{55} &\ding{55}&\ding{55}&\ding{51}\\
    \hline
    Top-1 Acc. (\%)& \baseline{69.90}  & 71.06 & 71.33 & {\bf71.96}\\
    Top-5 Acc. (\%)& \baseline{89.43}  & 89.51 & 90.25 & {\bf90.48}\\
    \shline
  \end{tabular}
   \end{center}
  \caption{Effects of our normalized non-target loss on ImageNet dataset. The teacher is ResNet34 and the student is ResNet18.}
  \label{table:ablation study}
\end{table}

\subsection{Difference between Our NKD and DKD}

To better use the soft labels, we decompose KD loss and normalize the non-target logits for a better performance. The decoupled method is inspired by DKD~\cite{zhao2022decoupled}. However, this paper presented a more straightforward and efficient decomposition method. DKD decouples KD loss as:
\begin{align}
\label{eq:dkd's kd}
    L_{kd}=&-T_{t}log (S_{t}) - (1-T_{t})log(1-S_{t}) \\
    &- (1-T_{t})\sum_{i \ne t}^{C}\hat T_{i}log(\hat S_{i})
\end{align}
\begin{align*}
    \hat T_{i} = \frac{T_{i}}{1-T_{t}},~~~\hat S_{i} = \frac{S_{i}}{1-S_{t}}.
\end{align*}
DKD analyzed the effects of KD's components and set hyper-parameters for a new formulation:
\begin{align}
\label{eq:dkd}
    L_{dkd}=&\alpha \cdot\big(-T_{t}log (S_{t}) - (1-T_{t})log(1-S_{t})\big) \\
    &- \beta\cdot\big(\sum_{i \ne t}^{C}\hat T_{i}log(\hat S_{i})\big)
\end{align}
While we decompose KD loss into the target loss (like the original CE loss) and non-target loss in CE form:
\begin{align}
    L_{kd}=-T_{t}log (S_{t}) - \sum_{i \ne t}^{C}T_{i}log(S_{i})
\end{align}
Then we find the sum of student's and teacher's non-target logits is different, making them hard to be the same. So we normalize them to equalize the sum as follows:
\begin{align}
    L_{nkd}=-T_{t}log (S_{t}) - \gamma\cdot\sum_{i \ne t}^{C}\mathcal{N}(T_{i})log(\mathcal{N}(S_{i}))
\end{align}
With this slight modification, we present our NKD loss and achieve better performance than DKD, as shown in Tab.~\ref{table:cifar results}.

\subsection{Effects of USKD's Target and Non-target Loss}
We propose a novel self-KD method called USKD, which utilizes customized soft labels that incorporate information from both target and non-target classes. To evaluate the impact of each type of information, we conduct experiments on MobileNet and RegNetX-1.6GF in Tab.~\ref{table:skd ablation}. The experimental findings demonstrate that both types of information are beneficial and important for the two models, and their combination leads to further improvements in performance. For instance, by combining the target and non-target class distillation together, MobileNet achieves 70.38\%, which surpasses the accuracy achieved by using distillation on either the target or non-target class alone.

\begin{table}[t]
  \begin{center}
  \setlength{\tabcolsep}{7 pt}
  \begin{tabular}{ccccc}
    \shline
    Loss & \multicolumn{4}{c}{Top-1 Accuracy (\%)}\\
    \hline
    Target  & \ding{55} &\ding{51}&\ding{55}&\ding{51}\\
    Non-target & \ding{55} &\ding{55}&\ding{51}&\ding{51}\\
    \hline
    MobileNet & \baseline{69.21} &70.18&69.43&{\bf70.38}\\
    RegNetX-1.6GF & \baseline{76.84} &76.87&77.25&{\bf77.30}\\
    \shline
  \end{tabular}
   \end{center}
  \caption{Ablation study of USKD's target and non-target loss. The experiments are conducted on the ImageNet dataset. All the results are the Top-1 accuracy (\%).}
  \label{table:skd ablation}
\end{table}

\subsection{Models with USKD for Downstream Task}
Our self-KD method yields remarkable accuracy gains for the classification task on CIFAR-100 and ImageNet datasets. To further evaluate its effectiveness and generalization, we also apply the pre-trained model to object detection using Mask R-CNN~\cite{he2017mask} as the detector and evaluate the model's performance with $AP^{box}$ and $AR^{box}$ on the COCO val2017 dataset~\cite{lin2014microsoft}. We conduct the detection experiments for 12 epochs using MMDetection~\cite{mmdetection}. As shown in Tab.~\ref{table:detection}, the ResNet-50 backbone trained with our method improves the detector's performance by 0.3 mAP and 0.5 mAR. The results demonstrate that our self-KD method not only improves the model's classification performance but also generalizes well to downstream tasks like object detection.

\begin{table}[t]
\setlength{\tabcolsep}{6 pt}
  \begin{center}
  \begin{tabular}{ccll}
    \shline
    \multirow{2}{*}{Method}&ImageNet&\multicolumn{2}{c}{COCO}\\
    &Top-1 Acc. (\%)&AP$^{box}$&AR$^{box}$\\
    \hline
    \baseline{Baseline (Res50)} &\baseline{76.55}&\baseline{38.0}&\baseline{52.4}\\
    {\bf USKD (ours)} &77.07&38.3 &52.9 \\
    \shline
  \end{tabular}
   \end{center}
    \caption{The detection results on the COCO dataset. We pre-train the backbone with USKD and use Mask-RCNN as the detector.}
  \label{table:detection}
\end{table}

\subsection{Customized Soft Labels for USKD}
\begin{figure}[t]
    \centering
    \includegraphics[width=1.0\linewidth]{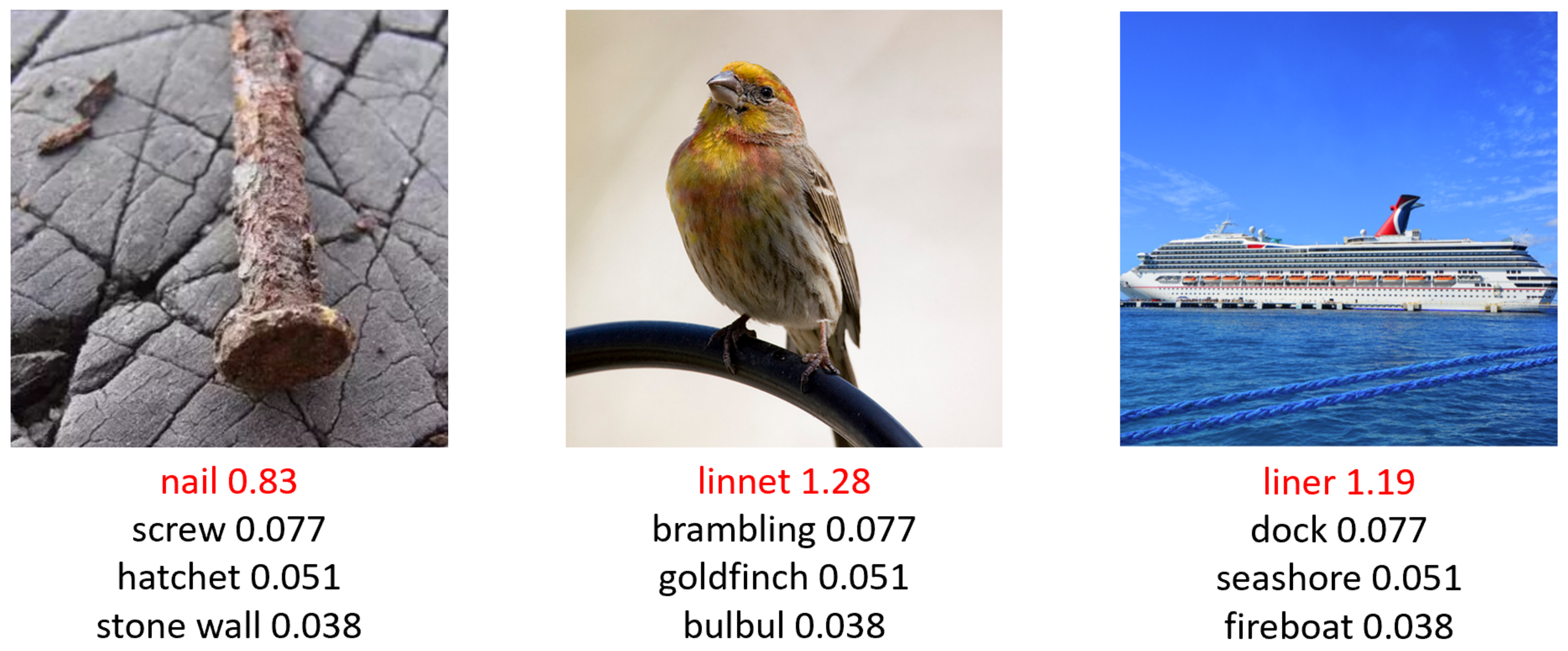}
    \caption{The visualization of the target and top-3 non-target class values of our customized soft labels. The target class value is obtained by squaring and smoothing. The non-target class value is obtained by Zipf's law.}
    \label{fig:softlabel}
\end{figure}

Our self-KD method, USKD, leverages customized soft labels for every image during training. Fig.\ref{fig:softlabel} shows several samples' soft labels, including the value for the target class and the top-3 non-target classes. For the target class, the value may be larger than 1 due to the smoothing method described in Eq.~\ref{eq:smooth}. For the non-target classes, USKD distributes larger values to the categories that are similar to the target class. For instance, the labels for the second image not only include the target `linnet,' but also assign higher values to similar non-target classes like `brambling', `goldfinch', and `bulbul'. This approach ensures that each image receives an appropriate customized soft label, which enables successful self-KD. The visualization also demonstrates how our USKD helps the model to perform better.

\subsection{Different Smooth Ways for Soft Target Label}
\label{sec:different smooth}
Our USKD utilizes customized soft labels that include target and non-target classes for all the images during training. To create the soft labels, we replace the weights of NKD's target loss with a smoothed version of the student's target output $S_t$, as shown in Eq.~\ref{eq:smooth}. In this subsection, we investigate the impact of different methods for smoothing the student's target output $S_t$ on the performance of the model. We conduct experiments on ResNet18 trained on ImageNet to compare these methods, as shown in Tab.~\ref{table:different ways}. The experiments show that all the methods significantly improve the student model. Notably, using $S_t+V_t-mean(S_t)$ as the weights for smoothing the student's target output achieves 0.86\% improvement, even outperforming using the teacher ResNet18's output as the soft target label. Based on this observation, we choose this method for smoothing the student's target output and use it as our soft target label.

\begin{table}[t]
  \begin{center}
  \begin{tabular}{lccc}
    \shline
    Smooth Ways &Teacher & Top-1 Acc. (\%) \\
    \hline
    \baseline{Baseline}       &\baseline{-}&\baseline{69.90}\\
    \hline
    $S_{t}+V_{t}-mean (S_{t})$     &\ding{55}&{\bf 70.76}\\
    $softmax (S_{t})*sum (V_{t})$   &\ding{55}&70.57 \\
    $\sqrt{S_{t}-min (S_{t})}$ &\ding{55}&70.57\\
    $S_{t}/max (S_{t})$        &\ding{55}&70.53 \\
    $S_{t}/mean (S_{t})$       &\ding{55}&70.50\\
    \hline
    Trained ResNet-18      &\ding{51}&{\bf 70.75}\\
    \shline
  \end{tabular}
     \end{center}
  \caption{Results of training ResNet18 with target loss on ImageNet dataset. All the operations are calculated with different samples in a training batch. Because the trained Res-18's output is $S_{t}$, we drop the square operation for different smoothing ways.}
  \label{table:different ways}
\end{table}

\subsection{Different Ranks for Soft Non-target Labels}
\label{sec:normal}
In Eq.~\ref{eq:non-target sort}, we combine the normalized weak and final logit to obtain the soft non-target labels' rank. In this subsection, we explore the effects of different ways for the rank, as shown in Tab.~\ref{table:sort}. Specifically, we compare the performance when using the weak logit's rank alone, the final logit's rank alone, or a combination of both but without normalization. Interestingly, our results show that all these methods help improve the model's accuracy, with the combination of the two normalized logits achieving the best performance. It is also noteworthy that even directly using the final logit to obtain the rank can satisfy accuracy gains. These findings provide insights into the effectiveness of various approaches for the rank of the non-target labels.

\begin{table}[t]
  \begin{center}
  \setlength{\tabcolsep}{10 pt}
  \begin{tabular}{lcc}
    \shline
     Model  &ResNet-18  &RegNetX-1.6GF\\
    \hline
    \baseline{Baseline}   &\baseline{69.90}&\baseline{76.84}\\
    \hline
    Weak Logit   &70.65&77.28\\
    Final Logit  &70.72&77.15\\
    Combination  &70.71&77.25\\
    Normalization  &70.79&77.30\\
    \shline
  \end{tabular}
     \end{center}
    \caption{Results of different ranks for soft non-target labels. \textbf{Normalization} and \textbf{Combination} mean combining weak and final logit with normalization and without normalization, respectively.}
  \label{table:sort}
\end{table}

\section{Conclusion}

In this paper, we decompose KD loss into the target loss like original CE loss and non-target loss in a CE form. We then normalize the non-target logits to enhance the student's learning from the teacher. With this normalization, we introduce Normalized KD (NKD), which helps students to achieve state-of-the-art performance. Building on our NKD loss formulation, we further propose a new self-KD method, USKD, that works for both CNN-like and ViT-like models. USKD uses customized soft labels that include target and non-target classes for self-KD. We first square and smooth the student's target output logit as the soft target label. For the soft non-target labels, we use weak supervision to obtain the rank and utilize Zipf's law to generate the labels. In this way, USKD needs negligible extra time and resources than training the model directly. Extensive experiments on various models with different datasets demonstrate that both our NKD and USKD are simple and efficient.

{\bf Acknowledgement.} This work was supported by the National Key R$\&$D Program of China (2022YFB4701400/4701402), the SZSTC Grant (JCYJ20190809172201639, WDZC20200820200655001), Shenzhen Key Laboratory (ZDSYS20210623092001004), and Beijing Key Lab of Networked Multimedia.

{\small
\bibliographystyle{ieee_fullname}
\bibliography{egbib}
}

\clearpage
\appendix
\section*{Appendix}
\section{Effects of the Temperature in NKD}
\label{sec:temp}
The temperature $\lambda$ in NKD is a hyper-parameter used to adjust the distribution of the teacher's soft non-target labels. NKD always applies $\lambda>1$ to make the logit become smoother, which causes the logit contains more non-target distribution knowledge. The target output probability of the same model will get a higher value on an easy dataset, such as CIFAR-100. This causes $ T_{i}^{\lambda}$ to contain less knowledge, which may bring adverse effects to the distillation. In this subsection, we explore the impact of using different temperatures to distill the student ResNet18 on CIFAR-100, shown in Tab.~\ref{table:ablation T}. The results show that temperature is an important hyper-parameter, especially for an easy dataset. For all the experiments, we adopt $\lambda=1$ on ImageNet. While for CIFAR-100, we follow the training setting from DKD for a fair comparison.

\section{NKD for ViT-liked Models}
\label{sec:vit-liked nkd}
To further evaluate the effectiveness of our NKD, we apply them to a vision transformer model DeiT, as shown in Tab.~\ref{table:deit results}. We conduct experiments on the tiny and base DeiT with NKD, bringing them excellent improvements. DeiT-Base achieves 84.96\% Top-1 accuracy with NKD, which is 3.20\% higher than the baseline. Besides, NKD also outperforms the classical KD for ViT's distillation, which proves the effectiveness of our modification of KD's formula again.

\section{Square for Soft Target Label}
\label{sec:square}
For the target loss $L_{target}$, we first square the student's target output and then smooth it as the soft target label. Here we explore the effects of the operation. We conduct experiments by training MobileNet and RegNetX-1.6GF on the ImageNet dataset, which is shown in Tab.~\ref{table:target}. The square enlarges the difference between different samples' $S_{t}$ in a training batch and brings more improvements to the model with self-knowledge distillation. Specifically, the model MobileNet achieves 70.18\% top-1 accuracy with our target distillation loss $L_{target}$. While the model's top-1 accuracy without square is 70.04\%. The results demonstrate the effectiveness of the square for the soft target label.

\section{Sensitivity Study of USKD's Parameters}
\label{sec:uskd sens}
\begin{figure}
  \begin{center}
    \includegraphics[width=0.9\linewidth]{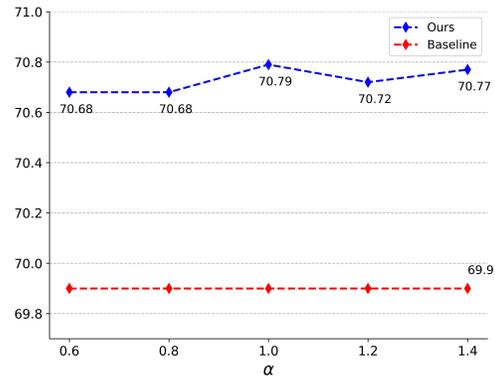}
    \end{center}
    \caption{The hyper-parameter $\alpha$ for target loss $\mathcal{L}_{target}$ with ResNet-18 on ImageNet dataset.}
    \label{fig:alpha}
\end{figure}

\begin{figure}
\begin{center}
    \includegraphics[width=0.9\linewidth]{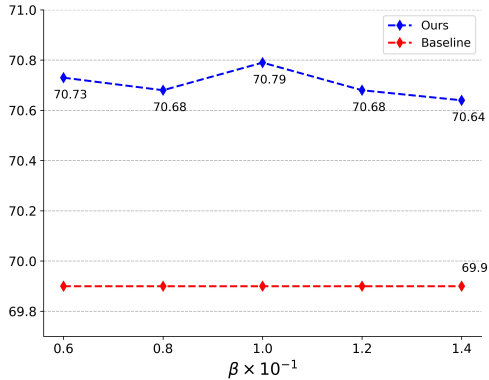}
    \end{center}
    \caption{The hyper-parameter $\beta$ for non-target loss $\mathcal{L}_{non}$ with ResNet-18 on ImageNet dataset.}
  \label{fig:beta}
\end{figure}

In our proposed method USKD, we use two hyper-parameters $\alpha$ and $\beta$ to balance the target loss $\mathcal{L}_{target}$ and the non-target loss $\mathcal{L}_{non}$, respectively. To explore the effects of the two hyper-parameters for self-KD, we conduct experiments by training ResNet-18 with our method on the ImageNet dataset. As shown in Fig.~\ref{fig:alpha}, $\alpha$ is used to adjust the target loss scale. Our method is not sensitive to it. When $\alpha$ varies from 0.6 to 1.4, the student's worst accuracy is $70.68\%$, which is just $0.13\%$ lower than the highest accuracy. Besides, it is still $0.78\%$ higher than training the model directly. As for the $\beta$ for the non-target loss in Fig.~\ref{fig:beta}, our method is not sensitive to it when $\beta < 0.1$. However, when it is too large, {\em e.g.} 0.14, the performance improvement may be affected.

\begin{table}[t]
\begin{center}
\begin{tabular}{c|c|cccc}
\shline
$\lambda$ &\baseline{Baseline}& 1.0 & 2.0 & 3.0 & 4.0\\
\hline
Top-1 &\baseline{78.58}& 80.55 & {\bf 80.76} & 80.72 & 80.54\\
Top-5 &\baseline{94.10}& {\bf 95.14} & {\bf 95.14} & 95.11 & 95.05\\
\shline
\end{tabular}
\end{center}
\caption{NKD's results on the CIFAR-100 with different temperature. We use ResNet-34 as the teacher to distill ResNet-18.}
\label{table:ablation T}
\end{table}

\begin{table}[t]
  \begin{center}
  \begin{tabular}{c|ll}
    \shline
    Teacher& Student  & Top-1 Acc. (\%)\\
    \hline
    \multirow{3}{*}{\makecell{DeiT III-Small\\(82.76)}}
    &DeiT-Tiny & \baseline{74.42}\\
    &KD        & 76.01 ({\color{red}+1.59}) \\
    &NKD (Ours)       & \textbf{76.68} ({\color{red}+2.26}) \\
    \hline
    \multirow{3}{*}{\makecell{DeiT III-Large\\(86.81)}}
    &DeiT-Base & \baseline{81.76}\\
    &KD        & 84.06 ({\color{red}+2.30})\\
    &NKD (Ours)       & \textbf{84.96} ({\color{red}+3.20})\\
    \shline
  \end{tabular}
  \end{center}
  \caption{Comparison of NKD and KD on DeiT on ImageNet-1k. The teacher is pre-trained on ImageNet-21K.}
  \label{table:deit results}
\end{table}

\begin{table}[t]
  \begin{center}
  \begin{tabular}{lcc}
    \shline
       &MobileNet  &RegNetX-1.6GF\\
    \hline
    \baseline{Baseline}   &\baseline{69.21}&\baseline{76.84}\\
    \hline
    w/o square   &70.04&76.79\\
    with square  &70.18&76.87\\
    \shline
  \end{tabular}
  \end{center}
  \caption{Comparison of training the models with different distillation methods on the target class. The experiments are conducted on the ImageNet dataset.}
  \label{table:target}
\end{table}

\section{Weak Supervision for the Weak Logit}
\label{sec:weak}
\begin{figure}[t]
    \centering
    \includegraphics[width=0.9\linewidth]{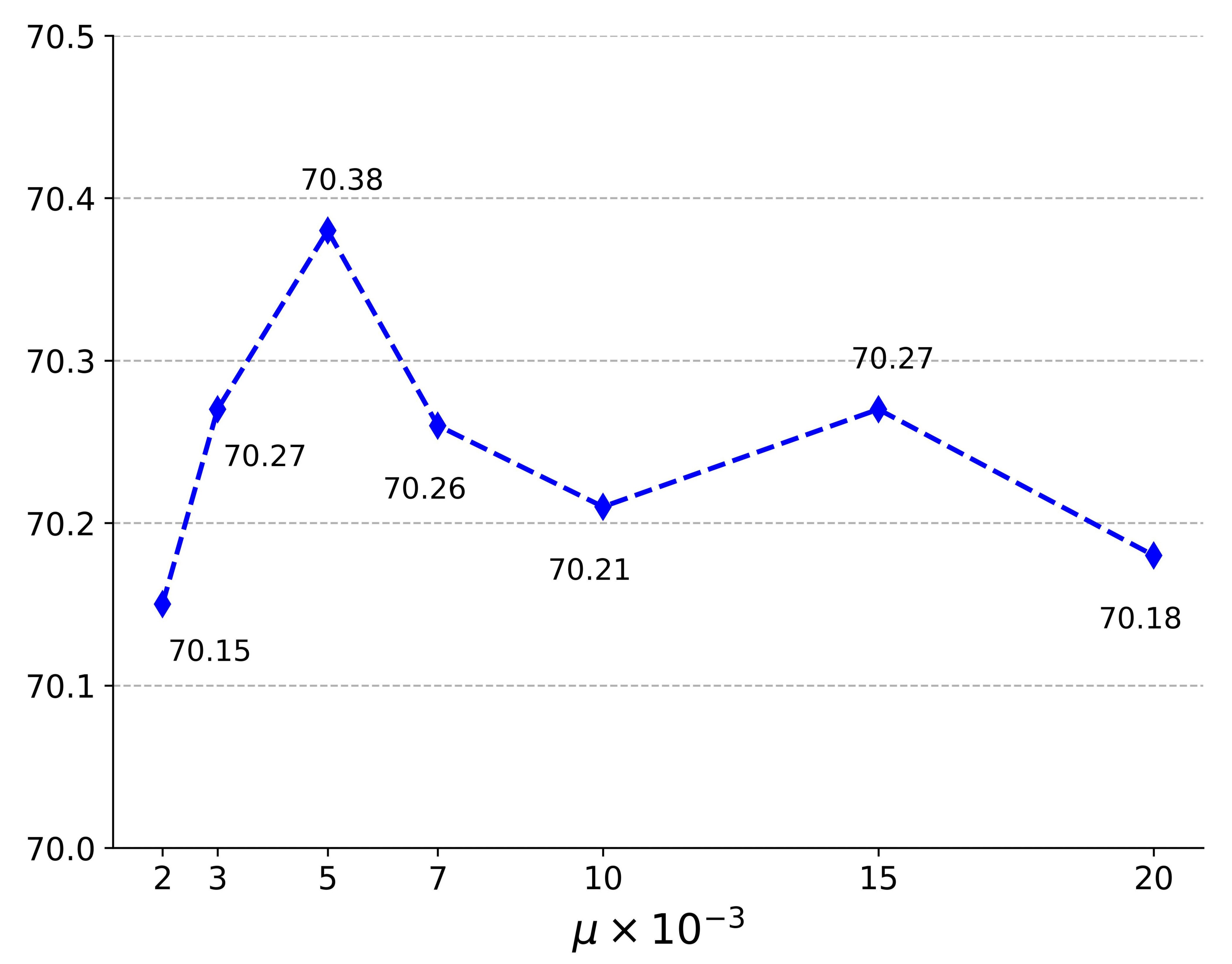}
    \caption{The hyper-parameter $\mu$ for weak supervision with RegNetX-1.6GF on ImageNet dataset.}
    \label{fig:weak supervision}
\end{figure}

For soft non-target labels' rank, we use a hyper-parameter $\mu<1$ to achieve weak supervision for the weak logit. Here we conduct experiments to investigate the influence of weak supervision on self-KD. For normal supervision, $\mu$ should be set to 1. However, the rank of the weak logit is similar to that of the final logit when $\mu=1$. With a smaller $\mu$, the supervision becomes weaker, and the difference between the two logits becomes larger. As shown in Fig.~\ref{fig:weak supervision}, the model's top-1 accuracy is 70.18\% when $\mu=0.02$. The RegNetX-1.6GF model achieves better performance with weaker supervision when $\mu > 0.005$. However, when $\mu$ is too small, for example, 0.002, the supervision is too weak, resulting in the model's weak logits being the same for all non-target classes, which negatively affects performance improvement. 

\section{Extension to Regression Tasks}
\label{sec:reg}
We apply NKD to object detection, surpassing KD and DKD in Tab.~\ref{table:detection}. NKD shows significant mAP gains of 1.3 for Faster RCNN, indicating its potential for other tasks.

\begin{table}[h]
\renewcommand\tabcolsep{8pt}

  \begin{center}
  \begin{tabular}{c|cccc}
    \shline
    COCO&\baseline{baseline}&+KD &+DKD &+NKD$_{ours}$\\
    mAP&\baseline{37.4}& 37.8& 38.2& \textbf{38.7} \\
    \shline
  \end{tabular}
   \end{center}
   \caption{The detection results on COCO. Teacher: Faster RCNN ResNet-101 (2x). Student: Faster RCNN ResNet-50 (1x).}
  \label{table:detection}
\end{table}

\section{Analysis on the Coefficient.}
\label{sec:coe}
The difference between the formula of NKD and DKD is $-(1-T_{t})log(1-S_{t})$. While DKD incorporates this term, NKD excludes it. DKD combines it with $T_{t}log(S_{t})$ as a whole to analyze its effects, and we only utilize $T_{t}log(S_{t})$ for distillation. In other word, DKD's coefficient for $log(1-S_{t})$ is $-(1-T_{t})$ and NKD's coefficient is 0. We conduct experiments to analyze the effects in Tab.~\ref{table:coff} for KD and Tab.~\ref{table:selfkd} for self-KD. 
The settings are: \textbf{T1:} $-T_{t}log(S_{t})$, \textbf{T2:} $-(1-T_{t})log(1-S_{t})$. Our NKD approach employs \textbf{T1}, while DKD utilizes \textbf{T1+T2}.

\begin{table}[h]
\renewcommand\tabcolsep{8pt}

  \begin{center}

  \begin{tabular}{cccc}
    \shline
    \baseline{Baseline}   &T1 (NKD)$_{ours}$ &T2& T1+T2 (DKD)\\
    \hline
    \baseline{69.90}& \textbf{71.96}   &70.91&71.70\\
    \shline
  \end{tabular}
     \end{center}

    \caption{KD results of Res18 on ImageNet. \textbf{T1:} $-T_{t}log(S_{t})$, \textbf{T2:} $-(1-T_{t})log(1-S_{t})$. NKD uses \textbf{T1} and DKD uses \textbf{T1+T2}.}

  \label{table:coff}
\end{table}

\begin{table}[h]
\vspace{-8pt}
  \begin{center}
  \small
  \begin{tabular}{cccc|c}
    \shline
    \baseline{Baseline}   &T1 (S1)$_{ours}$ &T2 (S1)&T1+T2 (S1)& T1 (S2)\\
    \hline
    \baseline{69.90}& \textbf{70.79} &69.50&70.22 &70.36\\
    \shline
  \end{tabular}
     \end{center}
    \vspace{-6pt}
    \caption{Self-KD results of Res18 on ImageNet. \textbf{S1} (USKD) and \textbf{S2} (EMA) denote different soft labels.} 
    \vspace{-6pt}
  \label{table:selfkd}
\end{table}

\end{document}